\documentclass[conference,a4paper,10pt]{IEEEtran}
\IEEEoverridecommandlockouts
\usepackage{cite}
\usepackage{amsmath,amssymb,amsfonts}
\usepackage{algorithmic}
\usepackage{graphicx}
\usepackage{textcomp}
\usepackage[dvipsnames]{xcolor}
\def\BibTeX{{\rm B\kern-.05em{\sc i\kern-.025em b}\kern-.08em
    T\kern-.1667em\lower.7ex\hbox{E}\kern-.125emX}}

\usepackage{amsmath,amssymb,amsfonts, enumitem, fancyhdr, color, comment, graphicx, environ,changepage, mathrsfs}
\usepackage{tikz}
\usepackage{tikz-cd}
\usepackage[inkscapelatex=false]{svg}
\usepackage{caption}
\usepackage{subcaption}
\usepackage{multicol}
\usepackage{url}

\def\gp{\mathcal{GP}}
\def\mean{\mathbb{E}}
\def\var{\text{Var}}

\def\data{\mathcal{D}}
\def\bI{\mathbf{I}}
\def\bK{\mathbf{K}}
\def\bk{\mathbf{k}}

\def\bx{\mathbf{x}}
\def\by{\mathbf{y}}

\begin{document}

\title{Trustworthy Prediction with Gaussian Process Knowledge Scores \\
\thanks{This work was supported by NSF under Award 2212506. We also acknowledge support of the UKRI AI programme, and the Engineering and Physical Sciences Research Council, for CHAI - Causality in Healthcare AI Hub [grant number EP/Y028856/1].}
}

\author{\IEEEauthorblockN{Kurt Butler,$^\dagger$ Guanchao Feng,$^\ddagger$ Tong Chen,$^\ddagger$ and Petar M. Djuri\'{c}$^\ddagger$}
\IEEEauthorblockA{
$^\dagger$\textit{School of Engineering},
\textit{The University of Edinburgh},
Edinburgh, UK \\
$^\ddagger$\textit{Department of Electrical and Computer Engineering},
\textit{Stony Brook University},
Stony Brook, NY, USA \\
\texttt{kbutler2@ed.ac.uk}, \texttt{\{guanchao.feng, tong.chen, petar.djuric\}@stonybrook.edu}}
}

\maketitle

\begin{abstract}
Probabilistic models are often used to make predictions in regions of the data space where no observations are available, but it is not always clear whether such predictions are well-informed by previously seen data. In this paper, we propose a knowledge score for predictions from Gaussian process regression (GPR) models that quantifies the extent to which observing data have reduced our uncertainty about a prediction. The knowledge score is interpretable and naturally bounded between 0 and 1. We demonstrate in several experiments that the knowledge score can  anticipate when predictions from a GPR model are accurate, and that this anticipation improves performance in tasks such as anomaly detection, extrapolation, and missing data imputation.
Source code for this project is available online at \url{https://github.com/KurtButler/GP-knowledge}.
\end{abstract}


\section{Introduction}
\begin{figure}[t]
    \centering
    \includegraphics[width=0.95\linewidth]{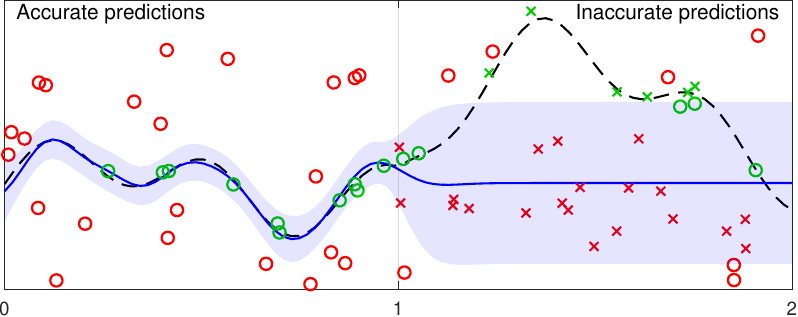}
    % \includesvg[width=0.95\linewidth]{Figures/cartoon.svg}
    \caption{Anomaly detection with a probabilistic model of the form $y=F(x)+noise$. The model is trained on a set of training data for values $x \leq 1$ (not pictured). The symbol $\times$ represents a point incorrectly identified as an anomaly or normal, and $\circ$ represents a correct assessment. \textcolor{Green}{Normal points} and \textcolor{Red}{anomalies} are shown in green and red respectively. The model is well-informed when $x \leq 1$, and the model becomes less useful for $x >1$ since it lacks data in this region. The problem is to determine the boundary between regions using the model alone. }
    \label{fig:cartoon}
    \vspace{-0.5cm}
\end{figure}

The task of prediction is of fundamental importance in many domains. Regression models are designed to predict the values of a function at new locations. In anomaly detection \cite{wang2019outlier, siffer2017anomaly}, a predictive distribution is used to decide if a newly received set of data contains abnormal samples. In data interpolation problems \cite{cesarelli2007algorithm}, the goal is to predict data that have been censored, corrupted, or are otherwise missing. While methods that rely on the ideas of prediction are widely deployed, they all require the common assumption that the predictive model is well-informed about what it intended to predict. Although common measures of accuracy on training or testing datasets help build confidence in the model, they fail to address a fundamental issue: the model's predictive reliability depends on where a given input lies within the input space, as shown in Fig. \ref{fig:cartoon}. This is evident in extrapolation problems, where the predictive confidence of the model decays as one strays farther out of distribution from the training data.

Recent work has noted that while a classification or regression method can produce a prediction, such predictions are not always well informed by training data \cite{jiang2018trust} and there may be reason for a model to abstain from producing an output \cite{dubuisson1993statistical, cordella1995adaptive}. Extrapolation problems \cite{wilson2013gaussian}, as noted above, represent one class of problems where this consideration occurs naturally. Another example is the problem of open-set recognition \cite{scheirer2012toward, mendes2017nearest}, where new samples that are far out-of-distribution (OOD) are suspected to be new, unseen classes, and so the classifier should withhold assigning the label of a known class to these new samples. In general, there has been a growing sentiment that machine learning models should know the limits of their own predictive knowledge \cite{ruggieri2025things}. These works highlight the growing interest in quantifying the trustworthiness of a model's prediction.

Recent works have proposed various methods to quantify the trustworthiness of an OOD prediction. Uncertainty quantification frameworks such as conformal prediction \cite{zhou2025conformal} or Bayesian credible intervals \cite{gelman1995bayesian} can assess trustworthiness by the size of a predictive interval. Other approaches define mathematical scores that assess whether a classifier's prediction is OOD by examining ratios of distances between nearby classes  \cite{jiang2018trust, mendes2017nearest}. These approaches have advantages and are built upon meaningful principles, but they also have their own quirks and pitfalls. For example, approaches based on predictive intervals require the designer to define when an interval becomes ``too large'', which is not always straightforward when using arbitrary units. For other methods \cite{mendes2017nearest}, the spatial resolution of novelty detection might be sensitive to the overlap between clusters. With considerations like this in mind, it becomes reasonable to investigate alternative approaches to prediction with a reject option.

In this paper, we introduce the concept of knowledge score, which can be applied to any probabilistic machine learning (ML) model. This mathematical score is derived from uncertainty quantification and is bounded between zero and one under certain conditions. The knowledge score measures to what extent a prediction from a probabilistic ML model is informed by the training data set. When the knowledge score is high, it indicates that the ML model is confident about what behavior it expects at a given location. When the knowledge score is low, the  model can indicate that it is relying mostly on the prior distribution, suggesting that a prediction is largely uninformed by the previously seen data. 
In this way, knowledge scores can be a valuable tool to quantify how much trust we should place in the predictions of the ML model. 
In this paper, we address the knowledge score of Gaussian processes (GPs) in the context of Gaussian process regression (GPR).
Our experiments show that the knowledge score can be applied to many problems, including extrapolation, interpolation and anomaly detection.

The paper is organized as follows. In Section \ref{sec:pf}, we formulate the problem we address. Section \ref{sec:met} describes the methodology for estimating the reliability of a GPR prediction, based on the concept of a knowledge score. The following section presents experimental results that demonstrate the application of the knowledge score to synthetic and real data. Finally, Section \ref{sec:con} provides concluding remarks.

\section{Problem Formulation}
\label{sec:pf}
We begin by recalling the general framework of Gaussian process regression (GPR) as motivation for our proposed knowledge score. Consider a regression model of the form, 
\begin{align}
    \label{eq:datamodel}
    y &= F(\bx) + \varepsilon, \\
    \varepsilon &\sim \mathcal{N}(0,\sigma^2),
\end{align}
where $\bx \in \mathbb{R}^D, y\in \mathbb{R}$, $F$ is an unknown nonlinear function and $\varepsilon$ is white Gaussian noise with variance $\sigma^2$. To estimate $F$ from data, we first place a GP prior over functions,
\begin{align}
    \label{eq:GPprior} 
    F & \sim \gp(0,k),
\end{align}
where $k$ is a kernel or covariance function. We assume that we have previously seen a data set $\data$ of input-output pairs,
\begin{equation}
    \label{eq:dataset}
    \data = \left\{ (\bx_n,y_n) : n=1, ..., N \right\}.
\end{equation}
Given prior observations at the locations $\mathbf{x}_n$, our predictive task requires us to predict the distribution at a new location $\bx$.
By conditioning on $\data$, one can obtain a GP posterior distribution over $F$,
\begin{align}
    \label{eq:GPposterior} 
    F|\data  & \sim \gp(m_p, k_p), \\
    m_p(\bx) &= \bk(\bx)^\top (\bK + \sigma^2 \bI)^{-1} \by,  \\
    k_p(\bx,\bx') &= k(\bx,\bx') - \bk(\bx)^\top (\bK + \sigma^2 \bI)^{-1} \bk(\bx'). 
    \label{eq:GPpostvar} 
\end{align}
The matrices $\bk(\bx)$ and $\bK$ are defined by
\begin{equation}
    \bk(\bx) = \begin{bmatrix}
        k(\bx,\bx_1) \\\vdots \\ k(\bx,\bx_N)
    \end{bmatrix},
    \bK = \begin{bmatrix}
        k(\bx_1,\bx_1) & \cdots & k(\bx_N, \bx_1) \\
        \vdots & \ddots & \vdots \\ k(\bx_1,\bx_N) & \cdots & k(\bx_N,\bx_N)
    \end{bmatrix},
    \notag
\end{equation}
and $\mathbf{y} = [y_1, \cdots, y_N]^\top$.

This posterior distribution over $F$ also induces a corresponding distribution over observations $y=F(\bx) + \varepsilon$, which takes on the form
\begin{equation}
    \label{eq:posteriorpred}
    y_* | \bx_*, \data \sim \mathcal{N}( m_p(\bx_*), k_p(\bx_*,\bx_*) + \sigma^2),
\end{equation}
which is used to infer the parameters of the GP kernel by maximizing the likelihood of $\data$.

In \eqref{eq:GPpostvar}, it can be seen that the posterior variance is strictly upper bounded by the prior variance: 
\begin{equation}
    \label{eq:variance_reduction}
k(\bx,\bx) - k_p(\bx,\bx) = \bk(\bx)^\top (\bK + \sigma^2 \bI)^{-1} \bk(\bx) \geq 0.
\end{equation}
This observation leads to the development of the knowledge score in the following section.

\section{Methodology}
\label{sec:met}
We now formulate the problem of quantifying model knowledge as a comparison of the prior distribution $p(F|\bx)$ and the posterior $p(F|\bx,\data)$. 
% \subsection{Knowledge quantification}
We begin with a formal definition, to define the {\em knowledge score} in terms of a general principle,
\begin{equation}
    \label{eq:Gfunc}
    G(\bx,\data) \stackrel{\Delta}{=} \frac{\var(F(\bx)|\bx) - \var(F(\bx)|\bx,\data)}{\var( F(\bx)| \bx)}.
\end{equation}
The function $G(\bx,\data)$ quantifies to what extent conditioning on the data set $\data$ has reduced the variance of the prediction $F(\mathbf{x})$ at the location $\bx$. Thus, we might also refer to $G$ as a variance reduction score.\footnote{We focus on the variance reduction score in this work. Entropy-based measures, as alternative measures, are left for future work.} A key insight of our approach is that the trustworthiness of a prediction depends critically on where the new location is in the input space with respect to previously seen data.

In the case of a GPR model, the function $G$ in \eqref{eq:Gfunc} becomes highly interpretable.
From the definition and \eqref{eq:variance_reduction}, it can be seen that $G$ is bounded between 0 and 1 for GPR models. A value near 0 indicates that the GP is not knowledgeable at a given location, in the sense that the prior and posterior distributions nearly coincide. If $G$ is near 1, then it indicates that conditioning on $\data$ has greatly reduced our uncertainty about $F(\bx)$.
Since the posterior variances in GPR have closed-form expressions, \eqref{eq:GPpostvar}, we see that $G$ can be expressed as
\begin{equation}
    \label{eq:Gfunc2}
    G(\bx,\data) = \frac{\bk(\bx)^\top( \bK + \sigma^2 \bI)^{-1} \bk(\bx) }{k(0,0)}.
\end{equation}

From \eqref{eq:Gfunc2}, it becomes apparent that besides $\data$ and $\bx$, the choice of kernel function $k$ also affects the knowledge score.  
Simple kernel functions such as the radial basis function (RBF) kernel \cite{rasmussen2006gaussian} are directly tied to the distances of each $\bx_n$ in $\data$ to the new location $\bx_*$. Learning of the hyperparameters of the kernel could then be tied to metric learning \cite{chen2024openset}. Other kernels, such as periodic or linear kernels, or combinations of kernel functions, can exhibit more complicated behavior that lead to more surprising patterns in the knowledge score. 
For example, if a kernel used in time series modeling is composed of a combination of a periodic and RBF kernel, then we might be interested in measuring the confidence of a forecast future data. The GP knowledge score provides a principled way to understand how the combination of these two kernels allows us to measure the confidence of our predictions as we extrapolate forwards in time. 

We now discuss properties of the knowledge score, and its applications to problems such as extrapolation, interpolation and anomaly detection.

\subsection{Properties of the Knowledge Score}
Several properties of the knowledge score are worth mention. Firstly, the score is naturally normalized and it is defined by an interpretable quantity (variance reduction).  Also, knowledge scores are agnostic about the distribution of the model inputs, $p(\bx)$ and the knowledge score does not require an output sample $y_*$ to be observed when making predictions about $F(\bx_*)$. From \eqref{eq:Gfunc}, it can be seen that the knowledge score is invariant under normalization of the output variable.

A minor limitation is that the knowledge score is only bounded between 0 and 1 for Gaussian models. In general, it is possible that $G<0$, which would indicate that observing $\data$ has \textit{increased} the uncertainty of a prediction. In this case, $G$ still provides useful information, but the interpretation may be different. In this work, we only constrain ourselves solely to GPR models.

\subsection{Improving Anomaly Detection with Knowledge Scores}
\label{sec:two_stage_AD}
GPs are natural tools for anomaly detection \cite{chandola2011gaussian, yang2024sequential, kowalska2012maritime}. Knowledge scores can be useful to withhold classifying new data when the GP model is uninformed by the training data.

To understand how knowledge scores can help with this task, we recall a typical formulation of the anomaly detection problem in the GPR framework.
Given the posterior predictive distribution in \eqref{eq:posteriorpred}, one can test if a novel observation pair $(\bx_*,y_*)$ behaves consistently with the previously observed data. If $y_*$ is far outside the distribution $p(y_*|\bx,\data)$, then the new pair $(\bx_*,y_*)$  may be marked as an outlier, because it is inconsistent with the previously observed data. 

To formalize the detection of outliers, we use the following decision rule that exploits Gaussianity of the posterior predictive distribution. We declare a new pair $(\bx_*,y_*)$ to be an anomaly if 
\begin{equation}
    \label{eq:anom}
    |y_* - \mean( F(\bx_*)|\bx_*,\data)| > \theta(\bx_*,\data),
\end{equation}
where the expectation is computed w.r.t. $F \sim p(F|\data)$, and $\theta$ is a threshold determined by some criterion. One common heuristic is to use 
$$
\theta(\bx_*,\data) = 3 \sqrt{ \var( y_*| \bx_*,\data) },
$$
which asserts that $y_*$ is an anomaly if the probability of more extreme value was very low (less than 0.3\%). 
Other principles may be used to select $\theta$, although for Gaussian distributions other principles typically yield tests of the form in \eqref{eq:anom}.

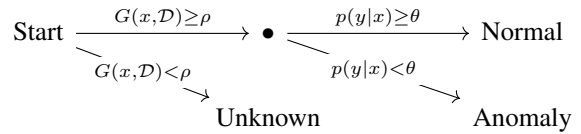
\begin{figure}
    \centering
    % https://q.uiver.app/#q=WzAsNSxbMCwwLCJcXHRleHR7U3RhcnR9Il0sWzIsMCwiXFxidWxsZXQiXSxbMiwxLCJcXHRleHR7VW5rbm93bn0iXSxbNCwxLCJcXHRleHR7QW5vbWFseX0iXSxbNCwwLCJcXHRleHR7Tm9ybWFsfSJdLFswLDEsIkcoeCxcXG1hdGhjYWx7RH0pXFxnZXEgciJdLFswLDIsIkcoeCxcXG1hdGhjYWx7RH0pIDwgciIsMV0sWzEsNCwicCh5fHgpXFxnZXFcXHRoZXRhIl0sWzEsMywicCh5fHgpXFxnZXFcXHRoZXRhIiwxXV0=
\[\begin{tikzcd}
	{\text{Start}} && \bullet && {\text{Normal}} \\
	&& {\text{Unknown}} && {\text{Anomaly}}
	\arrow["{G(x,\mathcal{D})\geq \rho}", from=1-1, to=1-3]
	\arrow["{G(x,\mathcal{D}) < \rho}"{description}, from=1-1, to=2-3]
	\arrow["{p(y|x)\geq\theta}", from=1-3, to=1-5]
	\arrow["{p(y|x)<\theta}"{description}, from=1-3, to=2-5]
\end{tikzcd}\]
    \caption{Flowchart describing a two-stage anomaly detection scheme using the  knowledge score. The symbols $\rho$ and $\theta$ represent thresholds.}
    \label{fig:two_stage_AD}
\end{figure}

For a new sample $(\bx_*,y_*)$, one can detect if the pair is anomalous using an anomaly detection approach such as the one above. However, implicit in this approach is the assumption that $p(y_*|\bx_*,\data)$ is well-informed by the data set $\data$. Even if $\data$ contains many clean recordings of input-output pairs that allow us to learn a model with good predictive power on the training set, it is possible and sometimes inevitable that the model will face difficulties when extrapolating to new input locations $\bx_*$ that are unfamiliar with respect to the locations recorded in $\data$. 
One remedy to this problem is to use models which will increase the posterior predictive variance in such regions, but inflation of the predictive variance artificially can also inflate the type I error (miss rate) of the detector. 

An alternative solution is to refuse to classify points for which the model is not confident, using a two-stage classification approach. We propose one version of this in Fig. \ref{fig:two_stage_AD}, where the  knowledge score is used to decide when the class is unknown. This first stage consists of a comparison of the  knowledge score $G(\bx,\data)$ to a threshold parameter $\rho$, where points are classified as unknown when $G < \rho$. The second stage performs anomaly detection on the points for which $ G \geq \rho$. Our empirical results suggest that the two-stage approach can be useful to identify subsets of data where the anomaly detection is performed with high accuracy. The question of how to select $\rho$ is left to future work, although one simple approach is to select $\rho$ such that the empirical performance of the anomaly detector, on a given training set of data, satisfies some problem-specific criteria.

\subsection{Improving Extrapolation/Interpolation with Knowledge Scores}
Both extrapolation and interpolation are examples of problems where we wish to estimate the value of a function at a location where we have not received data. In the extrapolation problem, we want to predict the values at locations which are out-of-distribution from the distribution of the input points, $p(\bx)$ \cite{wilson2013gaussian}. 

In  interpolation problems, there are locations where we have missing data where we seek to infer the value of the function \cite{cesarelli2007algorithm}. The missing data segments in an interpolation problem can have variable length, but typically there are observed data on either side of the missing data segment. When missing data segment is short, many different methods may be used to interpolate the function. As the segment of missing data becomes larger, it becomes less clear that a model is well-equipped to interpolate the data. Thus, a critical issue is to determine how long is too long to perform interpolation. 

\section{Experiments}
\label{sec;exp}
In this section, we demonstrate how GP knowledge scores can be applied to several problems of common interest. 
% Namely, we apply GP knowledge scores to anomaly detection, extrapolation and missing data interpolation. 

\begin{figure}[!t]
    \centering
    % \includesvg[width=0.9\linewidth]{Figures/toy.svg}
    \includegraphics[width=0.9\linewidth]{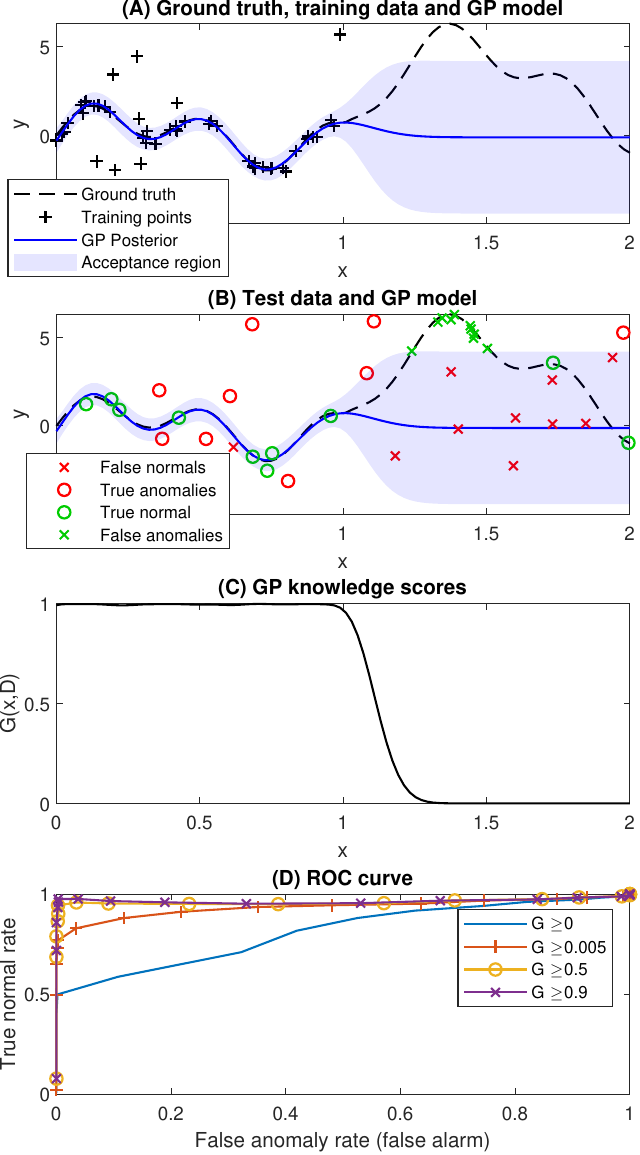}
    \caption{ 
    \textbf{(A)} Ground truth function, and training data used to train the GPR model for anomaly detection.
    \textbf{(B)} The GPR model and a subset of the test set predictions. 
    \textbf{(C)} The GP knowledge score $G(x,\data)$ viewed as a function of the input location $x$. 
    \textbf{(D)} Receiver-operating characteristic (ROC) curve showing the performance of the anomaly detector on subsets of the data. If we restrict ourselves to only analyzing points $x_n$ where $G(x_n,\data)$ exceeds some nonzero threshold, we observe that the performance of the model increases.
    }
    \label{fig:anomaly}
\end{figure}

\subsection{Trustworthy Anomaly Detection on Toy Data}
In this experiment, we consider how knowledge scores can be used to improve anomaly detection. Data are generated according to a model of the form \eqref{eq:datamodel}, where $x_n$ is randomly sampled from the range $0$ to $1$. Anomalies were generated according to a large variance distribution, $\mathcal{U}(-5,5)$, independently, for a small randomly selected minority of points. In total, 50 points were used for training. The training data was split into normal and anomaly classes using the GP mixture model mentioned previously. The performance of the GPR model for anomaly detection was then evaluated on 1000 test points. The utility of the knowledge score can be visualized in Fig. \ref{fig:anomaly}.

In the region $x>1$, the posterior predictive variance of the GPR grows rapidly, indicating that the model is not informed about the test points in that region. 
The knowledge score quantifies the growth of the variance in this region, by examining how much the posterior and prior variance differ. 
In the two-stage anomaly detection scheme of Sec. \ref{sec:two_stage_AD}, we propose that the knowledge score can classify the points as ``unknown'' when the predictive model is not confident enough to classify them as anomalous or normal. The performance of the anomaly detector on the remaining points should then improve. In Fig. \ref{fig:anomaly}, we observe that the receiver-operating characteristic (ROC) curves change depending on whether points are included or excluded due to their knowledge scores. As a threshold for $G$ is used to filter out points (i.e., when $\rho > 0$ in the two-stage anomaly detector), we observe that the accuracy of the anomaly detector improves considerably. As a result, the knowledge scores allow us to identify which portions of the test set are likely to be more accurate, even before we observe the test label values.

\subsection{Trustworthy Extrapolation on Electricity Data}
\begin{figure}
    \centering
    % \includegraphics[width=0.95\linewidth]{Figures/electricity_forecast.png}
    % \includesvg[width=0.95\linewidth]{Figures/extrapolation.svg}
    \includegraphics[width=0.95\linewidth]{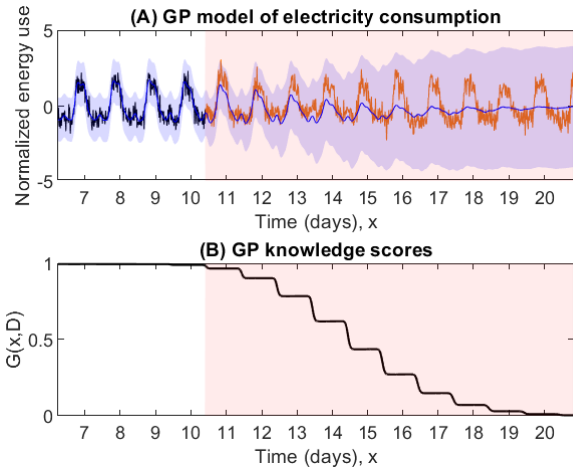}
    \caption{
    \textbf{(A)} Prediction of average consumption using a GPR model. Black points are observed real data, red points are unobserved (censored) real data, and the blue represents the GPR model. 
    \textbf{(B)} GP knowledge scores associated with the predictions. As time increases beyond the point of the data cutoff, the GP knowledge about predictions decreases.
    }
    \label{fig:electricity}
\end{figure}

In this experiment, we consider the problem of extrapolation using a real data set. We considered the Electricity Load Diagrams data set from the UCI Machine Learning Repository \cite{electricityloaddiagrams20112014_321}, which records the electricity consumption of 370 households in a Portuguese city over several years. With these data, we averaged and normalized the energy consumption across households to yield the energy load pattern for an average house in the city. A GPR model of energy consumption as a function of time was trained using the first 1,000 samples of data. We used a locally periodic kernel \cite{duvenaud2014automatic} for the GPR model, given by the formula,
\begin{align}
    k_{\text{LocPer}}(t,t') &= \sigma^2_f e^{\left(  \frac{ -2 \sin(\pi\rho^{-1} |t-t'| )}{\ell^2} \right)}
    e^{\left( -\frac{(t-t')^2}{2\ell^2} \right)},
    \notag
\end{align}
where $\sigma^2_f, \rho,$ and $\ell$ are hyperparameters.
The model was then tasked with extrapolating its predictions several days into the future, as shown in Fig. \ref{fig:electricity}.

It can be observed in the figure that the quality of the prediction degrades as the distance in time from the training data increases. As the distance increases, the posterior predictive distribution of the GP reverts to the prior. The GP knowledge scores track the transition back to the mean distribution. After many days, the predictive variance becomes large enough that it lacks descriptive power, indicating that the extrapolation scheme is only appropriate for a few days. Determining the exact period of time for which extrapolation is useful is a heuristic choice, but the GP knowledge score provides a clear and grounded metric for determining when the predictions have become too vague to be useful.

\subsection{Trustworthy Interpolation on \textcolor{black}{Fetal Heart Rate Recordings}}
In this experiment, we consider the problem of missing data interpolation using an open-access cardiotocography (CTG) data set \cite{chudavcek2014open} \textcolor{black}{in which 552 cardiotocography (CTG) recordings were selected from 9164 recordings acquired between April 2010 and August 2012 at the obstetrics ward of the University Hospital in Brno, Czech Republic. In CTG recordings, both fetal heart rate (FHR) and uterine activity (UA) measurements are usually obtained externally using ultrasound. As a result, various reasons, such as fetal or maternal movements and misplaced electrodes, can cause missing samples for varying periods of time.
Since computerized FHR analysis often relies on features extracted from FHR recordings, and many popular FHR features are sensitive to missing samples \cite{spilka2012stability}, properly addressing these missing FHR segments is of great importance. In practice, small segments of missing samples are interpolated using linear or cubic spline interpolation, while larger segments are often completely removed \cite{feng2017recovery}. However, there is no consensus on the missing segment length threshold for interpolation. It is worth noting that, similar to missing FHR segments, unreliably recovered/estimated FHR segments can also introduce serious distortions to downstream tasks and analyses. In this section, we show that trust score can be adopted to determine whether or not one should impute missing FHR segments in a principled manner. }

\begin{figure}
    \centering
    % \includesvg[width=0.85\linewidth]{Figures/fhr_one_col.svg}
    % \includesvg[width=0.95\linewidth]{Figures/interpolation.svg}
    \includegraphics[width=0.95\linewidth]{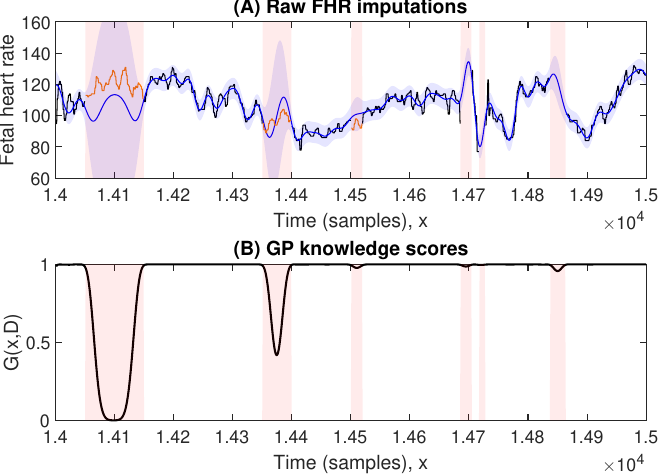}
    \caption{
    \textbf{(A)} Missing data interpolation with GPR. The red shaded regions represent missing data segments. Some data were manually removed (the red curves) to obtain a ground truth for some missing data segments. 
    \textbf{(B)} GP knowledge scores associated with GPR data interpolations. The dip in the GP knowledge score across missing data segments can be useful as a measure of where a missing data segment is too long to warrant interpolation.
    }
    \label{fig:fhr}
\end{figure}

In Fig. \ref{fig:fhr}, we visualize a GPR model used to interpolate missing samples from a randomly selected FHR recording. There are several missing segments of varying length in this example. Longer segments are less reliably interpolated, so a principled metric for determining when a missing data segment should be interpolated is desirable. We see that the GP knowledge score shows a clear distinction in how much the knowledge drops based on the length of the segment. As such, if the GP model is taken to be accurate, then those segments which exhibit only small decreases in knowledge scores can be safely interpolated. 
% Longer segments can be removed from interpolation, which reduces the chance of introducing artifacts that distort the downstream analysis. 

\section{Conclusion}
\label{sec:con}
In this paper, we introduced the concept of GP knowledge score to quantify the extent to which the GPR posterior distribution is informed by the training data. We apply the knowledge score to a few problems of interest and show that it can be a powerful tool to understand how models generalize.
In future work, we plan to study the relationship between knowledge scores and the empirical performance on specific tasks. We also consider information theoretic alternatives to our proposed variance reduction score. Finally, we consider relaxation of the Gaussianity assumption.

% \newpage

% Bibliography
\bibliographystyle{IEEEbib}
\bibliography{refs.bib}

\end{document}